\documentclass{article}

\usepackage{PRIMEarxiv}

\usepackage[utf8]{inputenc} 
\usepackage[T1]{fontenc}    
\usepackage{hyperref}       
\usepackage{url}            
\usepackage{booktabs}       
\usepackage{amsmath,amsfonts}       
\usepackage{nicefrac}       
\usepackage{microtype}      
\usepackage{lipsum}
\usepackage{fancyhdr}       
\usepackage{graphicx}       
\graphicspath{{media/}}     

\usepackage{bm}
\usepackage{physics}

\title{Fourier Analysis on Robustness of
Graph Convolutional Neural Networks for Skeleton-based Action
Recognition}

\author{
  Nariki Tanaka, Hiroshi Kera, Kazuhiko Kawamoto \\
  Chiba University, Japan \\
}

\begin{document}
\maketitle

\begin{abstract}
Using Fourier analysis, we explore the robustness and vulnerability of graph convolutional neural networks (GCNs) for skeleton-based action recognition. We adopt a joint Fourier transform (JFT), a combination of the graph Fourier transform (GFT) and the discrete Fourier transform (DFT), to examine the robustness of adversarially-trained GCNs against adversarial attacks and common corruptions. Experimental results with the NTU RGB+D dataset reveal that adversarial training does not introduce a robustness trade-off between adversarial attacks and low-frequency perturbations, which typically occurs during image classification based on convolutional neural networks. This finding indicates that adversarial training is a practical approach to enhancing robustness
against adversarial attacks and common corruptions in skeleton-based action recognition. Furthermore, we find that the Fourier approach cannot explain vulnerability against skeletal part occlusion corruption, which highlights its limitations. These findings extend our understanding of the robustness of GCNs, potentially guiding the development of more robust learning methods for skeleton-based action recognition.
\end{abstract}


\section{Introduction}\label{sec:introduction}
In skeleton-based action recognition, graph convolutional neural networks (GCNs) exhibit remarkable performance due to their ability to represent skeletal motion inputs using topological graphs~\cite{yan2018spatial,zhang2020semantics,cheng2020skeleton,chen2021channel,wang2022skeleton,lee2022hierarchically}.
However, recent studies have revealed that GCNs are
vulnerable to adversarial
attacks~\cite{diao2021basar, wang2021understanding, tanaka2022adversarial, liu2022adversarial}
and common corruptions, such as Gaussian noise.
This finding emphasizes the need to ensure
robustness in real-world applications.
To address these issues, other recent studies have proposed methods to improve the robustness of GCNs~\cite{yang2020improving,song2021richly,xing2022improved,wang2022defending_AAAI, shi2023occlusion}.
These vulnerabilities also imply that GCNs learn different features from humans, highlighting the need for a deeper understanding of their properties to develop robust models.

Recent studies in image classification have uncovered interesting properties of convolutional neural networks (CNNs) using Fourier analysis~\cite{Xu_ICONIP_2019, yin2019fourier,wang2020high,bernhard2021impact,abello2021dissecting,chan2022how,zhuang2022range}.
For example, \cite{yin2019fourier} discovered that adversarial perturbations for standard-trained CNNs are concentrated in the high-frequency domain, whereas those for adversarially-trained CNNs are concentrated in the low-frequency domain.
They also revealed that adversarial training encourages CNNs to capture low-frequency features of images, resulting in a trade-off between
robustness to low-frequency and high-frequency perturbations.
More specifically, adversarial training can enhance robustness to high-frequency corruptions, such as
Gaussian noise, while degrading robustness to low-frequency corruptions, such as fog corruption.
\cite{saikia2021improving} leveraged the trade-off to 
improve robustness by combining separate models robust
to low- and high-frequency perturbations.
Furthermore, \cite{zhuang2022range} observed that 
robust CNNs against common corruptions rely more on low-frequency features than standard-trained models.
These studies demonstrated the effectiveness of
Fourier analysis in understanding the robustness of CNNs.

\begin{figure}[t]
    \centering
    \includegraphics[width=0.8\columnwidth]{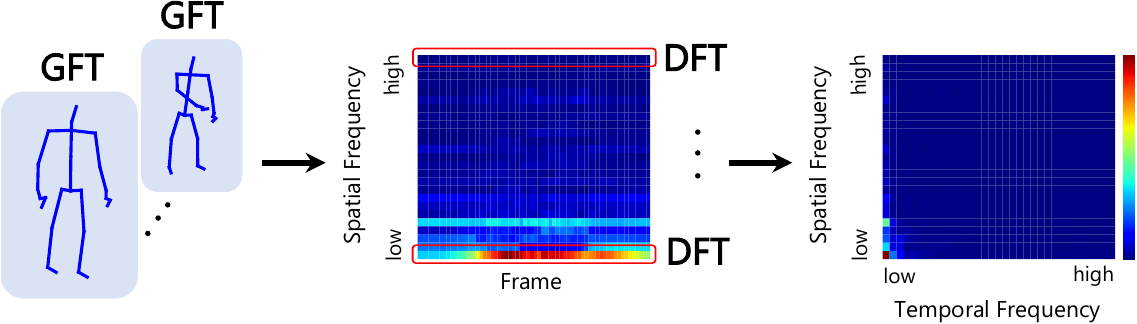}
\caption{Flow of the joint Fourier transform (JFT) on skeletal sequence data, which encompasses both the
graph Fourier transform (GFT) and the discrete Fourier
transform (DFT). The GFT is first applied to the skeletal data at each frame, followed by the DFT.}
\label{fig:intro}
\end{figure}

Inspired by these prior efforts, we examine the robustness of skeleton-based action recognition using Fourier analysis.
Unlike image classification, where the 2D discrete Fourier transform (DFT) can be used for Fourier analysis,
skeleton-based action recognition
requires an alternate approach owing to the graph-based representation of the skeletal data.
Therefore, we adopt a joint Fourier transform (JFT)~\cite{loukas2016frequency} that combines the graph Fourier transform~(GFT) and DFT,
as shown in Fig.~\ref{fig:intro}.
By applying the GFT to each frame of the skeletal sequence data and then using the DFT to each graph-frequency component, we analyze the frequencies of the skeletal sequence data along the spatial (graph-frequency) and temporal (time-frequency) directions. This method enables us to explore
the difference between CNN-based image classification and
GCN-based skeletal action recognition from a Fourier
perspective.
In experiments with the NTU RGB+D dataset~\cite{shahroudy2016ntu}, we apply
Fourier analysis to compare standard-trained and adversarially-trained GCNs.
Our observations reveal that there is no robustness trade-off between adversarial attacks and low-frequency perturbations for GCN-based skeletal action
recognition.
This finding is interesting because such a trade-off
is typically observed in image classification~\cite{yin2019fourier,chan2022how}.
Furthermore, we explore the robustness against common corruptions, such as Gaussian noise and part occlusion,
and find that an experimental result for the case of part occlusion
cannot
be explained by Fourier analysis alone.

The contributions of this study are as follows.
\begin{itemize}
\item A novel application of
Fourier analysis is presented to GCN-based skeletal-based action recognition.
Specifically, for the first time, we analyze the frequency characteristics of adversarially-trained GCNs against adversarial attacks and common corruptions using
the joint Fourier transform.

\item Experimental findings indicate that
there is no robustness trade-off between adversarial attacks and low-frequency perturbations for skeleton-based action recognition,
which is unique in CNN-based image classification.

\item Challenges are revealed in comprehensively explaining the robustness of skeleton-based action recognition using Fourier analysis.
Specifically, Fourier analysis cannot explain vulnerability against part occlusion corruptions.
\end{itemize}

The remainder of this paper is organized as follows.
Section \ref{sec:related_work} presents related work on the robustness of
skeleton-based action recognition and Fourier analysis approaches
in deep learning.
Section \ref{sec:Fourier_Analysis} describes a Fourier analysis method for GCNs.
Section \ref{sec:experiment} presents the experimental results.

\section{Related work}\label{sec:related_work}
This section reviews related work on the robustness of skeleton-based action recognition and
the Fourier analysis on robustness of deep models against adversarial attacks and common corruptions.

\subsection{Robustness of Skeleton-based Action Recognition}
Skeletal motion data offer several advantages over RGB videos, such as robustness to changes in clothing and background and superior computational efficiency~\cite{feng2022comparative,sun2023human}.
Due to the nature of the skeletal structure, GCNs have been extensively used for skeleton-based action recognition~\cite{yan2018spatial,zhang2020semantics,cheng2020skeleton,chen2021channel,wang2022skeleton,lee2022hierarchically}.

Recent research has explored adversarial attacks~\cite{Goodfellow2015ICLR,madry2018towards} on GCNs for skeleton-based action recognition and revealing the vulnerabilities of GCNs~\cite{diao2021basar,wang2021understanding,tanaka2022adversarial,liu2022adversarial}.
\cite{liu2022adversarial} presented the first white-box adversarial attack on GCNs. Their method perturbs joint locations while preserving temporal coherence, spatial integrity, and anthropomorphic plausibility.
\cite{diao2021basar} proposed a guided manifold walk method to search for adversarial examples in the black-box setting.
They also considered the naturalness and imperceptibility of perturbed skeletons.
\cite{wang2021understanding} proposed
an attack method in both white- and black-box settings
and highlighted the importance of considering motion dynamics in analyzing imperceptible adversarial attacks on 3D skeletal motion.
\cite{tanaka2022adversarial}
indicated that an adversarial attack was possible by only changing the lengths of the bones.
All of these studies demonstrate that GCNs are vulnerable to imperceptible adversarial perturbations.

Contrary to these studies on attack, research on defense against adversarial attacks for skeleton-based action recognition
remains in its infancy.
\cite{wang2022defending_AAAI} proposed
a Bayesian defense framework based on adversarial training, which
is the most effective defense technique against adversarial attacks.
Their study demonstrated that adversarial training is
effective not only for CNN-based image classification,
but also for GCN-based skeletal action recognition.
Nevertheless, existing studies on
adversarial robustness are limited and not sufficiently comprehensive.

Even so, vulnerability to common corruptions has been explored in several studies.
For example,
robustness against
Gaussian noise~\cite{yang2020improving},
part occlusion~\cite{song2021richly,xing2022improved,shi2023occlusion}, frame occlusions and jittering noises~\cite{xing2022improved}
have been explored.
Further investigation into the robustness against both common corruptions and adversarial attacks are essential for enhancing the applicability of deep models in real-world scenarios.

\subsection{Fourier analysis of CNN-based image classification}
In image classification, recent studies have attempted to explain the robustness of CNNs using frequency analysis~\cite{Xu_ICONIP_2019, yin2019fourier,wang2020high,bernhard2021impact,abello2021dissecting,chan2022how,zhuang2022range}.
\cite{Xu_ICONIP_2019} discovered Frequency
Principle (F-Principle) that CNNs initially
capture dominant low-frequency components before
slowly addressing high-frequency ones.
\cite{yin2019fourier} demonstrated that standard-trained CNNs predominantly depend on high-frequency components for image classification, while adversarial training encourages CNNs to capture low-frequency components of images.
Their further investigation revealed that
adversarial training enhances
robustness against common high-frequency corruptions, such as Gaussian noise, while degrading robustness against common low-frequency
corruptions, such as fog corruption.
These findings indicate the existence of a trade-off between robustness against high-frequency and low-frequency perturbations.
\cite{bernhard2021impact} observed that the frequency characteristics
of adversarial robustness may depend
on the dataset.
\cite{abello2021dissecting} investigated the
high-frequency biases of standard-trained CNNs.
\cite{chan2022how} explored these trade-offs by directly changing the frequency profile of the models.
\cite{zhuang2022range} identified
harmful frequencies for robustness to common corruptions and proposed a method to ignore these harmful frequency components.

\cite{wang2020high} explored the relationship between the frequency spectrum of image data and the generalization behavior of CNNs.
Their findings emphasized that while CNNs efficiently capture the high-frequency components of images, this capability renders them vulnerable. 
Conversely,
the robustness of CNNs can be enhanced by ignoring high-frequency components. 
These observations suggest that high-frequency and low-frequency components
in images correspond to domain-specific and domain-invariant features, respectively. This insight has subsequently been leveraged in domain generalization 
\cite{XU_CVPR_2021,Lin_2023_CVPR}.

These studies indicate that it is beneficial to use Fourier analysis to understand the
robustness of CNNs.
However, the frequency analysis of GCN-based skeleton action recognition remains unexplored.
Inspired by the Fourier analysis of CNNs by \cite{yin2019fourier}, we adapt their methodology to GCNs. 
This analysis allows us to explore the frequency features captured by GCNs and to assess their robustness across different frequency bands. 
Furthermore, using the Fourier analysis, we examine the robustness of skeleton-specific common corruptions
such as frame loss. These investigations promise to shed light on developing robust GCN models for skeleton-based action recognition.

\section{Fourier Analysis for Skeleton-based Action Recognition}
\label{sec:Fourier_Analysis}
Here, we investigate the frequency responses
of GCNs for skeleton action recognition
to reveal their robustness against adversarial attacks and common corruptions.
Rather than using the 2D discrete Fourier transform (DFT),
we employ
the joint Fourier transform (JFT), which is
a frequency analysis tool for time-varying graph signals~\cite{loukas2016frequency},
Furthermore, we use the Fourier heatmap~\cite{yin2019fourier} to visualize the sensitivity of deep models in the frequency domain.

\subsection{Spatiotemporal Graph for Skeletal Sequence Data}\label{subsec:notations}
A skeleton sequence for a
single person is represented
by an undirected graph $G = \qty(V, E)$, where $V$ is a set of nodes and $E$ is a set of edges.
The node set $V$ consists of all $N$ skeleton joints
, i.e., $V$ includes $N\times T$ nodes,
and the edge set $E$ indicates the intrabody 
connections of the skeleton.
Each node $v_i(t)\in V$ is associated with
a feature vector of the corresponding
joint position
$\bm{x}_{i}(t)=(x_{i}(t),y_{i}(t),z_{i}(t))^\top\in\mathbb{R}^3$ at frame $t$.
For example,
considering the joint positions themselves,
the node feature is $\bm{x}_{i}(t)$.
As with other features,
joint motion $\bm{v}_{i}(t)=\bm{x}_{i}(t+1)-\bm{x}_{i}(t)$,
bone $\bm{b}_{i}(t)=\bm{x}_{i}(t)-\bm{x}_{j}(t)$
($v_i$ and $v_j$ is a pair of connected joints),
and bone motion $\bm{v}_{i}^{\mathrm{b}}(t)=\bm{b}_{i}(t+1)-\bm{b}_{i}(t)$ have been used~~\cite{cheng2020skeleton,chen2021channel,wang2022skeleton}.
The set of such feature vectors is represented by
\begin{align}
\bm{X}=
\begin{pmatrix}
f_{1}(1) & f_{1}(2) & \ldots f_{1}(T)\\
f_{2}(1) & f_{2}(2) & \ldots f_{2}(T)\\
\vdots & \ddots & \vdots\\
f_{N}(1) & f_{N}(2) & \ldots f_{N}(T)\\
\end{pmatrix}\in\mathbb{R}^{N\times T},
\label{equ:skeletal feature}
\end{align}
where $f_{i}(t)$ is an
$x$, $y$, or $z$ coordinate
element of the feature vector associated with $v_i(t)$.
We omit the dimension of the three-axis channels for simplicity, although the related skeleton sequence is
represented by an $N \times T \times 3$ tensor.
The skeleton sequence is processed independently for each channel.

To normalize the frame lengths of all skeletal data,
linear interpolation along the temporal direction
are usually used~\cite{chen2021channel,wang2022skeleton, lee2022hierarchically}.
We denote the interpolated data for $\bm{X}$
by $I(\bm{X})\in\mathbb{R}^{N \times T^{\prime}}$, where $T^{\prime}$ is set to $64$ in our experiments.

\subsection{Standard \& Adversarial Training}\label{subsec:standard_training}
Let $\mathcal{M}_\theta$ be a GCN parameterized by $\theta$.
Standard training for $\mathcal{M}_\theta$ is performed by solving
the following minimization problem
\begin{align}
    \min_{\theta} \mathbb{E}_{\qty(\bm{X}, y) \sim \mathcal{D}} \qty[\mathcal{L}\qty(\mathcal{M}_\theta \qty(I(\bm{X}), y ))],
\end{align}
where $\mathcal{L}(\cdot)$ is the cross-entropy loss function
and $\mathcal{D}$ is an underlying data distribution
over pairs of interpolated skeletal data $I(\bm{X})$ and action labels $y$.

Adversarial training in this study is based on
the projected gradient descent (PGD)~\cite{madry2018towards},
which is a typical and strong attack method.
The PGD uses a set of perturbations $S=\{\bm{\delta} \mid \norm{\bm{\delta}}_p \leq \epsilon\}$, where
$\norm{\bm{\delta}}_p$ is the $l_{p}$ norm of $\bm{\delta}\in \mathbb{R}^{T\times N}$ ( we set $p$ as 2)
and $\epsilon > 0$ denote the supremum of the perturbation norm.
Adversarial training for $\mathcal{M}_\theta$ with the PGD is performed by solving
the following min-max optimization problem~\cite{madry2018towards}
\begin{align}
    \min_{\theta} \mathbb{E}_{\qty(\bm{X}, y) \sim \mathcal{D}} \qty[\max_{\bm{\delta}_\mathrm{adv} \in S} \mathcal{L}\qty(\mathcal{M}_\theta \qty( I(\bm{X}+\bm{\delta}_\mathrm{adv}), y ))],
\end{align}
where linear interpolation $I(\cdot)$ is repeatedly applied to adversarially perturbed data $\bm{X}+\bm{\delta}_\mathrm{adv}$ in the optimization loop.

\subsection{Discrete \& Graph Fourier Transforms}
\label{subsec: DFT_GFT}
DFT and GFT form the basis for the JFT on
spatiotemporal graph $G$,
as is discussed in the following subsection.
Additionally, DFT and GFT are used to generate
low-frequency or high-frequency Gaussian noise signals, which are used to validate the trade-off in robustness against low-frequency and high-frequency corruptions, as discussed in Section~\ref{sec:trade-off}.

We let the GFT be defined first.
The GFT on $\bm{X}$
is defined by
\begin{align}
\label{eq:gft}
\text{GFT}\qty(\bm{X})= \bm{U}^\top\bm{X},
\end{align}
where
$\bm{U}\in\mathbb{R}^{N\times N}$
is the eigenvector matrix of the Laplacian matrix $\bm{L}$
of the spatial subgraph of $G$.
The Laplacian matrix is
defined by $\bm{L} = \bm{D} - \bm{A}\in\mathbb{R}^{N\times N}$, where $\bm{A}$ and $\bm{D}$ denote the adjacency and degree matrices of the spatial skeletal structure, respectively.
Let $\lambda_k\in\mathbb{R}$ and $\bm{u}_k\in\mathbb{R}^N$ be the $k$-th eigenvalue and eigenvector
of $\bm{L}$.
The eigenvector matrix $\bm{U}$ is obtained by
applying the eigen-decomposition to $\bm{L}$ as follows
\begin{align}
\bm{L} = \bm{U}\bm{\Sigma} \bm{U}^\top,
\end{align}
where $\bm{U} = \qty[\bm{u}_1, \ldots, \bm{u}_N]$ and
$\bm{\Sigma} = \mathrm{diag}\qty(\lambda_1, \ldots, \lambda_N)$.
Here, the eigenvalues are sorted as $\lambda_1 \geq \lambda_2 \geq \cdots > \lambda_N=0$, where
the larger the eigenvalue is, the higher the frequency.

The DFT on $\bm{X}$
is defined by
\begin{align}
\label{eq:dft}
\text{DFT}\qty(\bm{X})=  \bm{X}
\bm{W},
\end{align}
where $\bm{W}\in\mathbb{R}^{T\times T}$ is the DFT matrix,
defined by
\begin{align}
\bm{W}=
\begin{pmatrix}
 1 & 1 & 1 & \ldots & 1\\
 1 & \omega & \omega^2 & \ldots & \omega^{T-1}\\
 1 & \omega^2 & \omega^4 & \ldots & \omega^{2(T-1)}\\
 \vdots & \vdots & \vdots & \ddots & \vdots\\
 1 & \omega^{T-1} & \omega^{2(T-1)} & \ldots & 
 \omega^{(T-1)(T-1)}\\
\end{pmatrix},
\end{align}
where $\omega=e^{-2\pi j/T}$ and $j^2=-1$.

\begin{figure}[t]
    \centering
    \includegraphics[width=0.8\columnwidth]{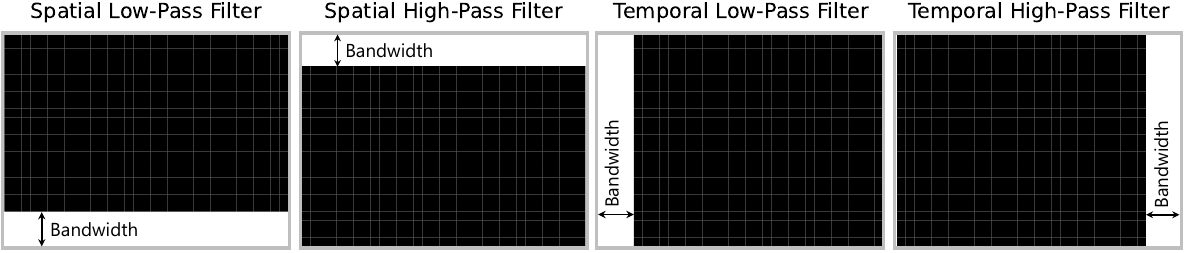}
\caption{Spatial low-pass (leftmost) and high-pass (left second) filtering are performed by masking
the Fourier spectrum along the spatial frequency axis.
Temporal low-pass (right second) and high-pass (rightmost) are performed by masking the Fourier spectrum along
the temporal frequency axis.}
    \label{fig:filter}
\end{figure}

\begin{figure}[t]
    \centering
    \includegraphics[width=0.8\columnwidth]{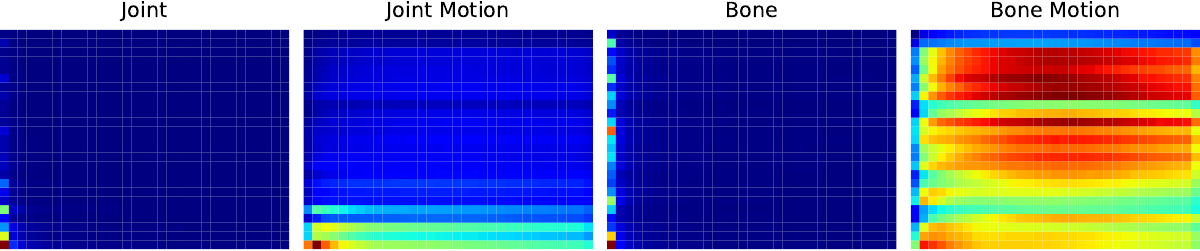}
\caption{Average Fourier spectrum over all tests
skeleton data for joint, joint motion,
bone, and bone motion features.
The vertical and horizontal axes represent spatial
and temporal frequency, respectively.
From these figures, for example,
we can see the Fourier spectrum of the joint feature
(leftmost)
is concentrated at low frequencies in the spatiotemporal frequency domain.}
    \label{fig:input_spectrum}
\end{figure}

We generate
low-frequency or high-frequency Gaussian noises that is 
added to skeletal data $\bm{X}$
using the GFT and DFT.
Let $\bm{V}$
be the Gaussian white noise on $\mathbb{R}^{N\times T}$,
where each element of $\bm{V}$ is independently sampled from
a zero-mean Gaussian distribution $\mathcal{N}(0,\sigma^2)$.
Spatial filtering or graph spectral filtering
on $\bm{V}$ is defined by
\begin{align}
    \widetilde{\bm{V}_\mathrm{s}}=
    \bm{M}_\mathrm{s}\text{GFT}(\bm{V}),
\end{align}
where $\widetilde{\bm{V}_\mathrm{s}}$ is the
filtering output in the spatial frequency
domain and
$\bm{M}_\mathrm{s}=\mathrm{diag}(m_{\mathrm{s},1},m_{\mathrm{s},2},\ldots,m_{\mathrm{s},N})$ is the $N\times N$
binary diagonal matrix for spatial filtering.
For instance,
if $\bm{M}_\mathrm{s}=\mathrm{diag}(0,0,\ldots,0,1,1)$, as shown in Fig.~\ref{fig:filter} (leftmost),
a low-frequency Gaussian noise with a bandwidth of 2 is generated by applying the inverse
GFT to $\widetilde{\bm{V}_\mathrm{s}}$.
If $\bm{M}_\mathrm{s}=\mathrm{diag}(1,1,0,\ldots,0)$, as shown in Fig.~\ref{fig:filter} (left second),
High-frequency Gaussian noise with a bandwidth of 2 is generated.
Similarly, temporal filtering on $\bm{V}$ is defined by
\begin{align}
    \widetilde{\bm{V}_\mathrm{t}}=\text{DFT}(\bm{V})
    \bm{M}_\mathrm{t},
\end{align}
where $\widetilde{\bm{V}_\mathrm{t}}$ is the
filtering output in the temporal frequency
domain and $\bm{M}_\mathrm{t}=\mathrm{diag}(m_{\mathrm{t},1},m_{\mathrm{t},2},\ldots,m_{\mathrm{t},T})$ is the $T\times T$
binary diagonal matrix for temporal filtering.
Similar to the GFT,
$\bm{M}_\mathrm{t}=\mathrm{diag}(0,0,\ldots,0,1,1)$  and
$\bm{M}_\mathrm{t}=\mathrm{diag}(1,1,0,\ldots,0)$
generate low-frequency and high-frequency Gaussian noises with a bandwidth of 2,
as shown in
Fig.~\ref{fig:filter} (right second)
and (rightmost), respectively.

\subsection{Joint Fourier Transform and Fourier Heatmap}
We adopt the JFT~\cite{loukas2016frequency} for Fourier analysis on spatiotemporal graph G, which encompasses both the GFT and DFT.
Initially, the GFT is implemented on the spatial subgraph of $G$ for each frame to extract spatial frequency characteristics, followed by the application of the DFT to extract temporal frequency characteristics of $G$,
as shown in Fig.~\ref{fig:intro}.

The JFT on $\bm{X}$ is defined by
combining the GFT and DFT as follows:
\begin{align}
\label{eq:jft}
\text{JFT}\qty(\bm{X})= \bm{U}^\top\bm{X}\bm{W}.
\end{align}
For Fourier analysis on skeletal data,
we compute the average spectrum over all test data of $\bm{X}$ using the JFT,
and visualize these values using the heatmaps shown in Fig.~\ref{fig:input_spectrum}, where 
the horizontal and vertical axes represent temporal and spatial frequencies, respectively.

Additionally, we employ Fourier heatmaps~\cite{yin2019fourier}
to visualize the sensitivity of the GCNs for specific frequency signals.
A signal of spatial frequency $\lambda_k$ (the $k$-th eigenvalue of $\bm{L}$) and temporal frequency $l/T$, denoted by $\bm{F}_{k,l}\in\mathbb{R}^{N\times T}$,
is generated by setting the JFT of $\bm{F}_{k,l}$ to
\begin{align}
    \text{JFT}(\bm{F}_{k,l})= 
    \bordermatrix{
    &        &   &   & l^\text{th} &  &  & \cr
    & 0      &  & 0 & 0 & 0 & \ldots & 0 \cr
   & \vdots & \vdots & \vdots & \vdots & \vdots & \vdots & \vdots \cr
  k^\text{th} & 0 & \ldots & 0 & 1 & 0 & \ldots & 0 \cr
    & \vdots & \vdots & \vdots & \vdots & \vdots & \vdots & \vdots \cr
    & 0      & \ldots & 0 & 0 & 0 & \ldots & 0 \cr
    },
\end{align}
where the right-hand side matrix
takes $1$ for only the $(k,l)$-element and $0$ otherwise.
Using $\bm{F}_{k,l}$,
a perturbed signal of a skeleton sequence $\bm{X}$
is generated by
\begin{align}
    \bm{X}'=I(\bm{X})+rv\bm{F}_{k,l},
    \label{equ:FHM}
\end{align}
where $r$ is sampled uniformly at random from $\qty{-1, 1}$,
$v>0$ is the norm of the perturbation,
and $\bm{F}_{k,l}$ is supposed to be normalized as $||\bm{F}_{k,l}||_2=1$.
The Fourier heatmap is generated by plotting
the average error rate of
1000 randomly sampled
test data for every frequency $k=1,2,\ldots,N$ and $l=1,2,\ldots,T$.
The closer to red a region of the Fourier heatmap is, the more sensitive the GCNs at the corresponding frequency. In other words, to recognize human actions, GCNs capture skeletal signals that contain such sensitive frequencies.
The high sensitivity brings vulnerability to the GCNs.

\begin{figure}[t]
    \centering
    \includegraphics[width=\linewidth]{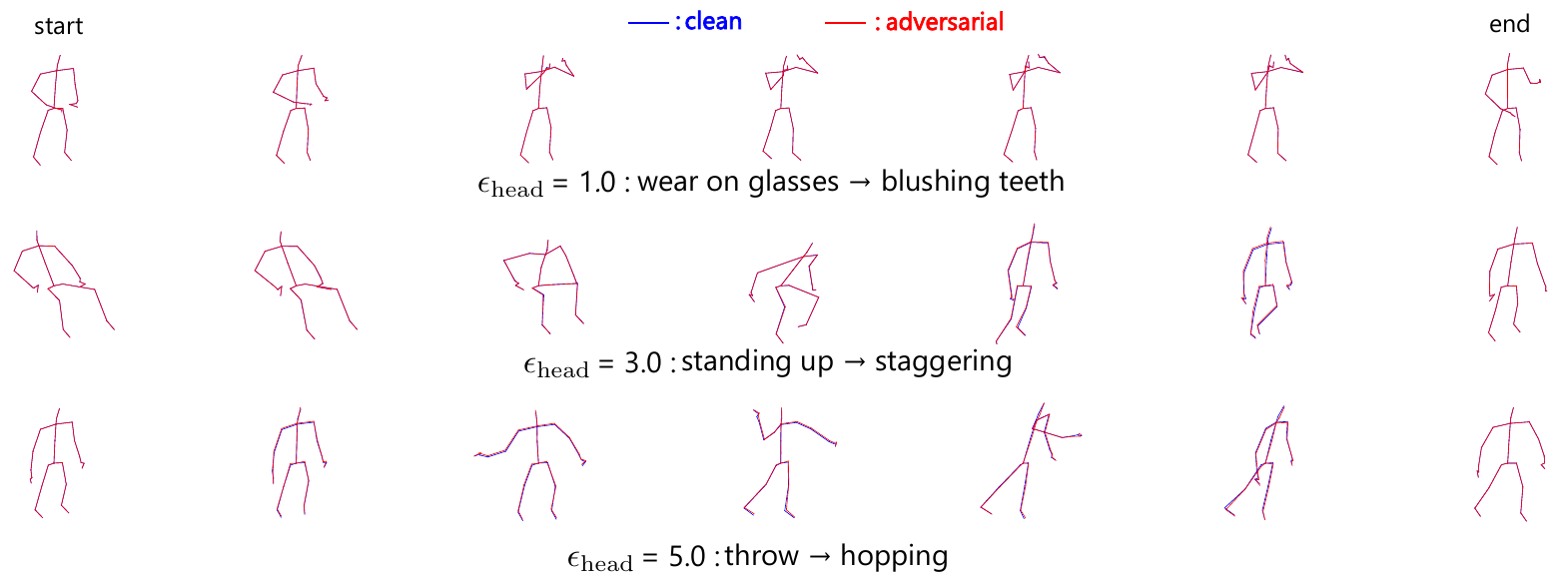}
\caption{Adversarial examples generated by the
$l_2$-PGD.
Clean (blue) and adversarial (red) examples are superimposed for three example actions.
These two almost overlap and are highly imperceptible. For each action, action labels before and after adversarial attacks are provided.}
    \label{fig:adversarial_examples}
\end{figure}

\section{Experiment}\label{sec:experiment}
Using frequency analysis, we conduct a comparative analysis of the robustness for the standard-trained and adversarially-trained GCNs.
First, we investigate the robustness and vulnerability
in the frequency domain using Fourier heatmaps.
The visualization provides basic insights into the robustness of adversarial training.
Next, we analyze the spectral distributions of adversarial perturbations to the GCNs.
This analysis reveals the frequency characteristics
of adversarial attacks.
Furthermore,
we explore whether the robustness trade-off that has been established in CNN-based image classification exists similarly in GCN-based skeletal action recognition.
Finally, we evaluate the robustness of the GCNs against
common corruptions.

\subsection{Experimental Setting}
\subsubsection{Dataset}
We conduct our experiments on NTU RGB+D~\cite{shahroudy2016ntu},
which contains 56,880 skeletal data with 60 action classes.
We divide the whole data into training and test data according to the subjects (cross-subject setting).
The training and test datasets comprise
40,320 and 16,560 samples, respectively.
For the validation data, we randomly sample 5\% from
the training data.

\subsubsection{Model}
We chose ST-GCN~\cite{yan2018spatial} as a baseline
GCN and TCA-GCN~\cite{wang2022skeleton} as one of the 
state-of-the-art GCNs.
We have not assessed robust models, 
as it is challenging to distinguish between GCN's inherent robustness and potential vulnerabilities against adversarial attacks and common corruptions. Our primary focus is on evaluating the pure robustness of GCN-based action recognition. 
A comprehensive validation against robust models will be addressed in future work.

We train ST-GCN and TCA-GCN for
joint $\bm{x}_i(t)$,
joint motion $\bm{v}_i(t)$, bone $\bm{b}_i(t)$, and bone motion
$\bm{v}_i^\mathrm{b}(t)$, respectively, using official codes provided by those authors.
To guarantee convergence in training, we execute at least 80 epochs for the ST-GCN and 75 epochs for the TCA-GCN and adopt early stopping with patience 20 for both models.
For the other hyperparameters, we use the same ones of the official codes.

\begin{table}[t]
  \caption{Comparison of clean accuracy between standard-trained (ST) and adversarially-trained (AT) models.}
  \centering
  \begin{tabular}{c|c|c|c|c} \hline
  Model & Joint & Joint Motion & Bone & Bone Motion \\ \hline
  ST-GCN (ST) & \textbf{85.3}\% & \textbf{84.8}\% & \textbf{85.4}\% & \textbf{84.4}\% \\ 
  ST-GCN (AT) &  79.8\% & 73.1\% & 79.7\% & 76.0\% \\ \hline
  TCA-GCN (ST) & \textbf{89.4}\% &\textbf{86.7}\% & \textbf{89.2}\% &\textbf{86.3}\% \\
  TCA-GCN (AT) & 82.1\% &  77.2\% &  82.2\% & 78.4\% \\ \hline
  \end{tabular}
  \label{tab:clean_acc}
\end{table}

\begin{table}[t]
  \centering
  \caption{Comparison of adversarial accuracy between standard-trained (ST) and adversarially-trained (AT) models.}
  \begin{tabular}{c|c|c|c} \hline
    Model & $\epsilon_\mathrm{head}= 1$ & $ \epsilon_\mathrm{head} = 3 $ & $ \epsilon_\mathrm{head} = 5$ \\ \hline
    ST-GCN ST (Joint)  & 26.3\% & 6.00\% & 1.63\%\\ 
    ST-GCN AT (Joint) & \textbf{74.6}\% & \textbf{61.2}\% & \textbf{47.7}\% \\ \hline
     TCA-GCN ST (Joint) & 16.7\% & 1.72\% & 0.19\%\\ 
    TCA-GCN AT (Joint) & \textbf{76.8}\% & \textbf{64.6}\% & \textbf{52.4}\% \\ \hline
    ST-GCN ST (Joint Motion) & 9.30\% & 0.61\% & 0.06\% \\ 
    ST-GCN AT (Joint Motion) & \textbf{64.2}\% & \textbf{48.3}\% & \textbf{35.8}\% \\ \hline
    TCA-GCN ST (Joint Motion) & 2.84\% & 0.04\% & 0.00\% \\ 
    TCA-GCN AT (Joint Motion) & \textbf{69.0}\% & \textbf{51.6}\% & \textbf{36.5}\% \\ \hline
    ST-GCN ST (Bone) & 8.99\% & 0.35\% & 0.07\% \\ 
    ST-GCN AT (Bone) & \textbf{71.5}\%& \textbf{52.5}\%& \textbf{32.9}\%\\ \hline
    TCA-GCN ST (Bone) & 8.35\% & 0.21\% & 0.01\% \\ 
    TCA-GCN AT (Bone) &\textbf{76.1}\% & \textbf{60.8}\% & \textbf{44.5}\% \\ \hline
    ST-GCN ST (Bone Motion) & 0.87\% & 0.02\% & 0.00\% \\ 
    ST-GCN AT (Bone Motion) & \textbf{62.2}\% & \textbf{40.5}\% & \textbf{24.5}\% \\ \hline
    TCA-GCN ST (Bone Motion) & 0.39\% & 0.03\% & 0.02\% \\ 
    TCA-GCN AT (Bone Motion) & \textbf{61.3}\% & \textbf{46.5}\% & \textbf{31.8}\% \\ \hline
  \end{tabular}
  \label{tab:attack}
\end{table}

\subsubsection{Adversarial Attack}
To evaluate adversarial robustness, we use the $l_2$-PGD attack rather than skeleton-specific attacks such as SMART \cite{wang2021understanding} and 
BASAR \cite{diao2021basar}.
These skeleton-specific attacks are designed to be imperceptible by imposing additional constraints 
such as fixing bone length.
These constraints complicate the direct control of the attack strength. In contrast, the PGD attack strength can be simply controlled by increasing a perturbation threshold $\epsilon$. Therefore, using the PGD attack, we can systematically assess the robustness of GCNs. 
Furthermore, we can compare GCN-based action recognition with CNN-based
image classification,
because the PGD attack is commonly used for CNNs
 (e.g., as seen in \cite{yin2019fourier}).

\begin{figure}[t]
\centering
\includegraphics[width=\linewidth]{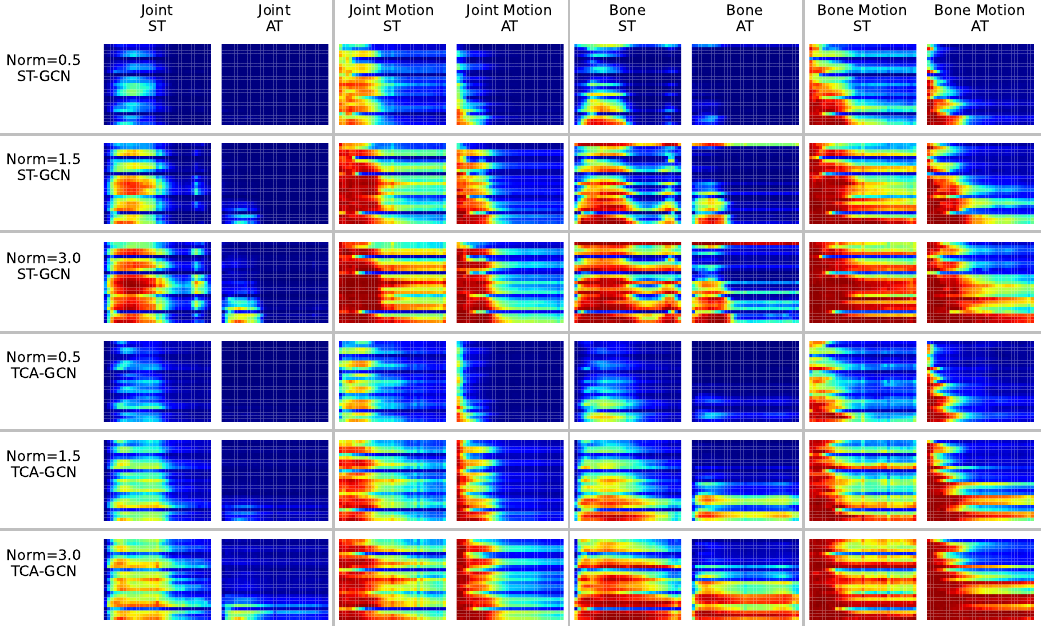}
\caption{Fourier heatmaps of standard-trained (ST) and adversarially-trained (AT) GCNs.
The top and bottom three rows display those of
the ST-GCNs and TCA-GCNs, respectively,
for each of the four features (joint, joint motion, bone, bone motion) and three perturbation norms $v\in\{0.5,1.5,3.0\}$.
}
\label{fig:heatmap}
\end{figure}

To normalize the attack strength to each piece of skeletal data,
we set the perturbation threshold $\epsilon$
as $\epsilon = l_\mathrm{head}\times \epsilon_\mathrm{head}$, where $l_\mathrm{head}$ is
the head length of each skeletal datum.
Fig.~\ref{fig:adversarial_examples} displays adversarial examples when $\epsilon_\mathrm{head}=1.0,3.0,5.0$.
We do not impose any naturalness constraints on the attacker, 
except for the perturbation norm.
Nevertheless, we observe that adversarial attacks are highly imperceptible because linear interpolation
in Section \ref{subsec:notations}
suppresses large jittering.

\subsubsection{Adversarial Training}
Adversarial training~\cite{madry2018towards} is an
effective method for defending against adversarial attacks.
We use Free~\cite{shafahi2019adversarial} for adversarial training due to its computational efficiency.
For Free, the number of hop steps is set to four, and
the threshold is set to $\epsilon_\mathrm{head}=3$.
Table~\ref{tab:clean_acc} shows the accuracies of standard-trained and adversarially-trained models. As is well established, adversarial training performs less effectively than standard training for clean data.
To verify that adversarial training enhances adversarial robustness, we conduct preliminary experiments on robustness
of the standard-trained and adversarially-trained models with thresholds $\epsilon_\mathrm{head} \in \qty{1.0, 3.0, 5.0}$ and 10 iterations.
Then, we attack each model by perturbing clean data correctly classified by both models to take the difference in clean accuracy into consideration.
In Table~\ref{tab:attack}, we can see that adversarial training demonstrates higher adversarial robustness than standard training.

\subsubsection{Evaluation Metric}
As shown in Table~\ref{tab:clean_acc}, there is a difference in clean accuracy between standard-trained and adversarially-trained models.
For a fair comparison of robustness, we remove a such difference.
More specifically, when we evaluate the robustness, we perturb clean data correctly classified by both models and use accuracy for these perturbed data as an evaluation metric.

\subsection{Results}

\subsubsection{Frequency Analysis of Adversarial Training}

\begin{figure}[t]
\centering
\includegraphics[width=0.8\columnwidth]{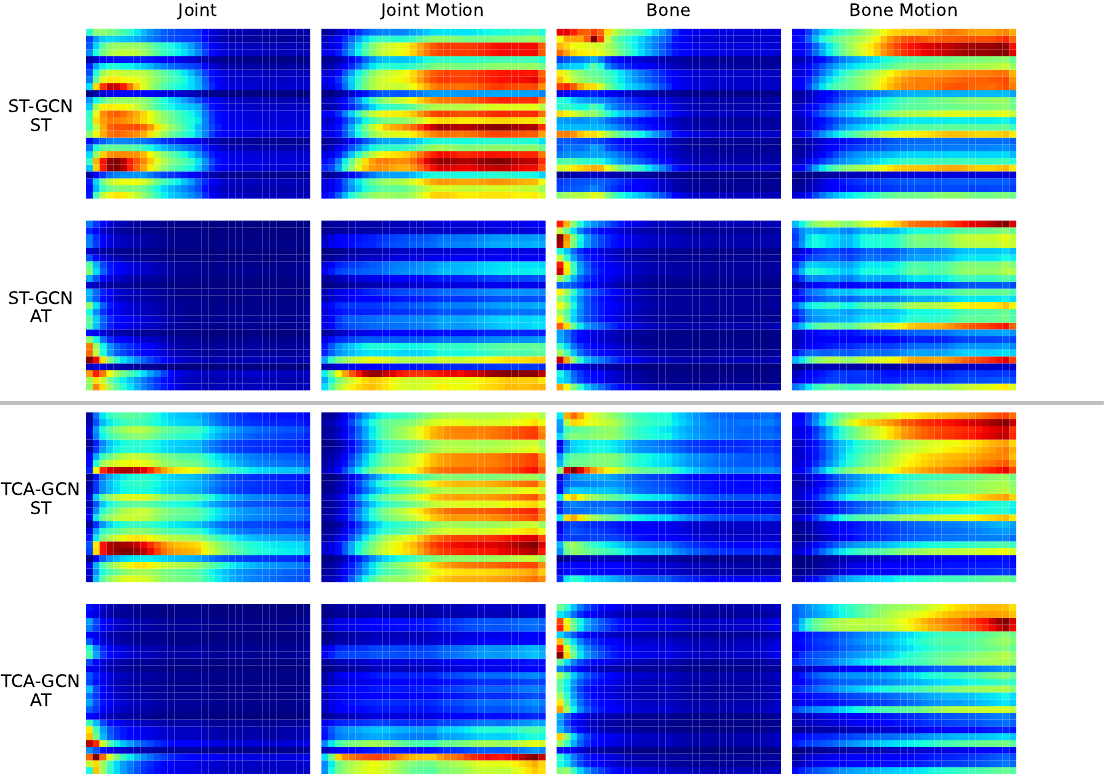}
\caption{Average Fourier spectrum of adversarial perturbations in the spatiotemporal frequency domain.
The heatmaps plot the estimation of
$\mathbb{E}\qty[|\mathrm{JFT}(I(\bm{X}+\bm{\delta}_\mathrm{adv})-I(\bm{X}))|]$,
where $\bm{X}$ is the clean data and the expectation is chosen over the adversarial examples that successfully
attack each model.}
\label{fig:perturbations}
\end{figure}

The Fourier heatmap examines the frequency characteristics of standard and adversarial training.
Fig.~\ref{fig:heatmap} shows
Fourier heatmaps plot the average error rates of 1000 randomly sampled test data points for every frequency.
The top and bottom three rows display
the Fourier heatmaps of the ST-GCNs and TCA-GCNs, respectively, for each of the four features (joint, joint motion, bone, bone motion) in Section~\ref{subsec:notations}
and three perturbation norms $v\in\{0.5,1.5,3.0\}$ in Eq.~(\ref{equ:FHM}).

Fig.~\ref{fig:heatmap} reveals the following frequency characteristics.
The standard-trained models are sensitive to
temporal low-frequency perturbations (i.e., the left half of the maps).
In other words, the models do not capture
high-frequency signals in time, whereas they do capture signals from low to high frequencies in space.
The adversarially-trained models become more insensitive,
especially for high-frequency perturbations (i.e., the
upper right of the maps),
resulting in robustness to high-frequency perturbations.
Joint motion and bone motion features
introduce more vulnerability against high-frequency perturbations than the
joint and bone features, respectively.
This vulnerability is reasonably attributed to motion (differential) features generally being more sensitive to high-frequency perturbations than positional features.
These frequency characteristics do not depend on the
network architecture, and the two GCNs have similar characteristics.

\begin{table}[t]
  \caption{List of scaled norms for spatially or temporally
  filtered Gaussian noises.}
  \centering
  \begin{tabular}{c|c|c|c|c} \hline
    Filter & Joint & Joint Motion & Bone & Bone Motion\\ \hline
    Spatial Low & 17.0 & 3.50 & 6.40 & 1.10 \\ \hline
    Spatial High & 77.0 & 20.0 & 5.90 & 4.80 \\ \hline
    Temporal Low & 31.0 & 2.0 & 9.90 & 0.60\\ \hline
    Temporal High & 77.0 & 19.0 & 24.0 & 14.0\\ \hline
  \end{tabular}
  \label{tab:scaled_norm1}
\end{table}

\begin{figure}[t]
\centering
\includegraphics[width=0.8\columnwidth]{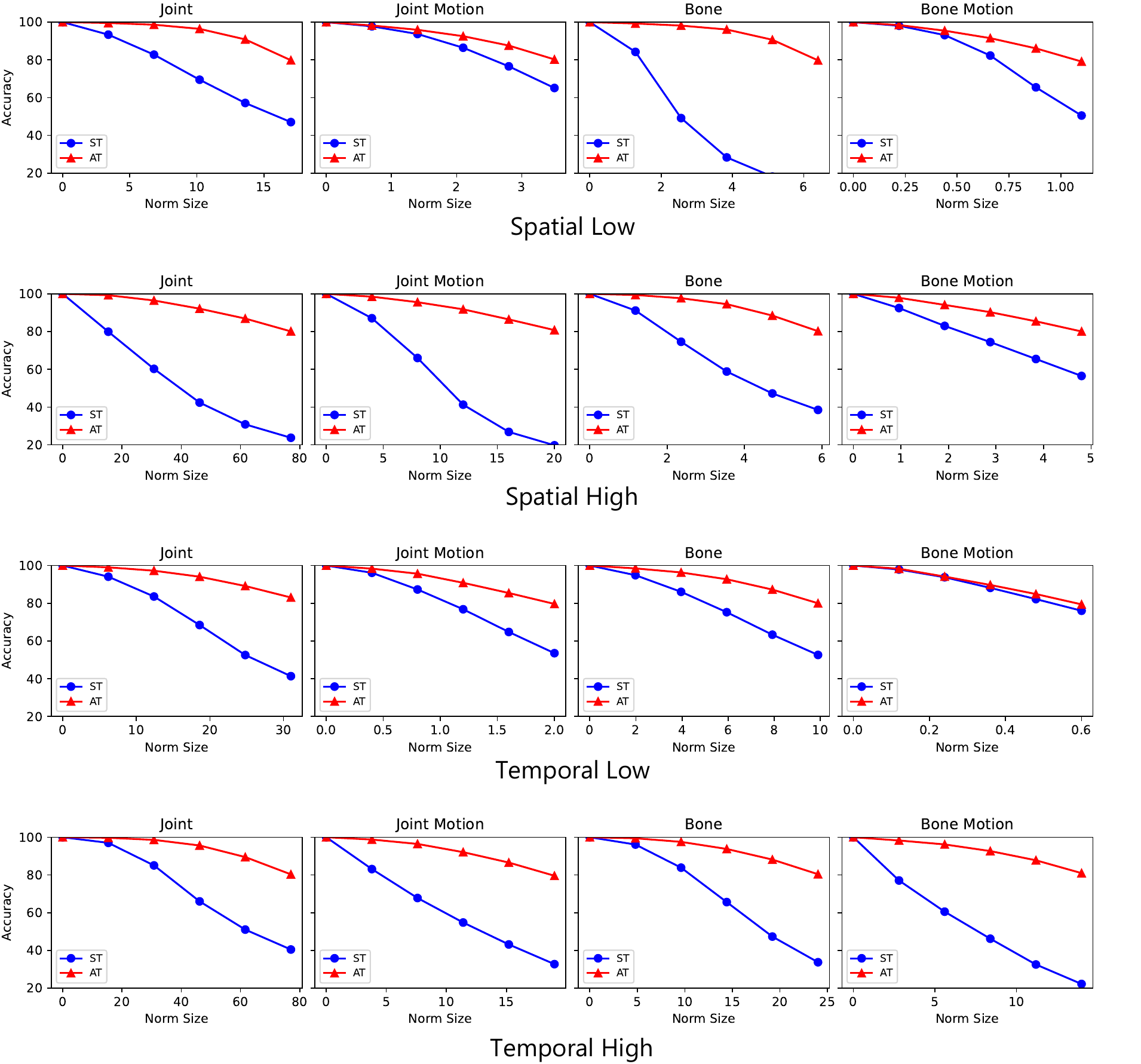}
\caption{Accuracy comparison of standard-trained (blue) and adversarially-trained (red) GCNs for changes in the
Gaussian noise norm with a fixed bandwidth of 2:
(top) spatial low-pass filter, (second) spatial high-pass
filter, (third) temporal low-pass filter, and (bottom)
temporal high-pass filter.}
\label{fig:gauss_norm}
\end{figure}

These results prove that adversarial training can improve
the robustness in the higher frequencies,
and the improvement is
also observable
in CNN-based image classification~\cite{yin2019fourier,wang2020high,abello2021dissecting}.
In contrast,
the lower frequency characteristics
of the CNNs and GCNs
are slightly different.
In CNN-based image classification, adversarial training sacrifices the robustness of the standard-trained models in the lowest frequencies (i.e., the lower left of the maps),
whereas this is not always the case
in GCN-based action recognition.
For example, the adversarially-trained models learned with the joint and bone features do not become more vulnerable than the standard-trained models
at the lowest frequencies.

\subsubsection{Frequency Analysis of Adversarial
Attack}

\begin{figure}[t]
\centering
\includegraphics[width=0.8\columnwidth]{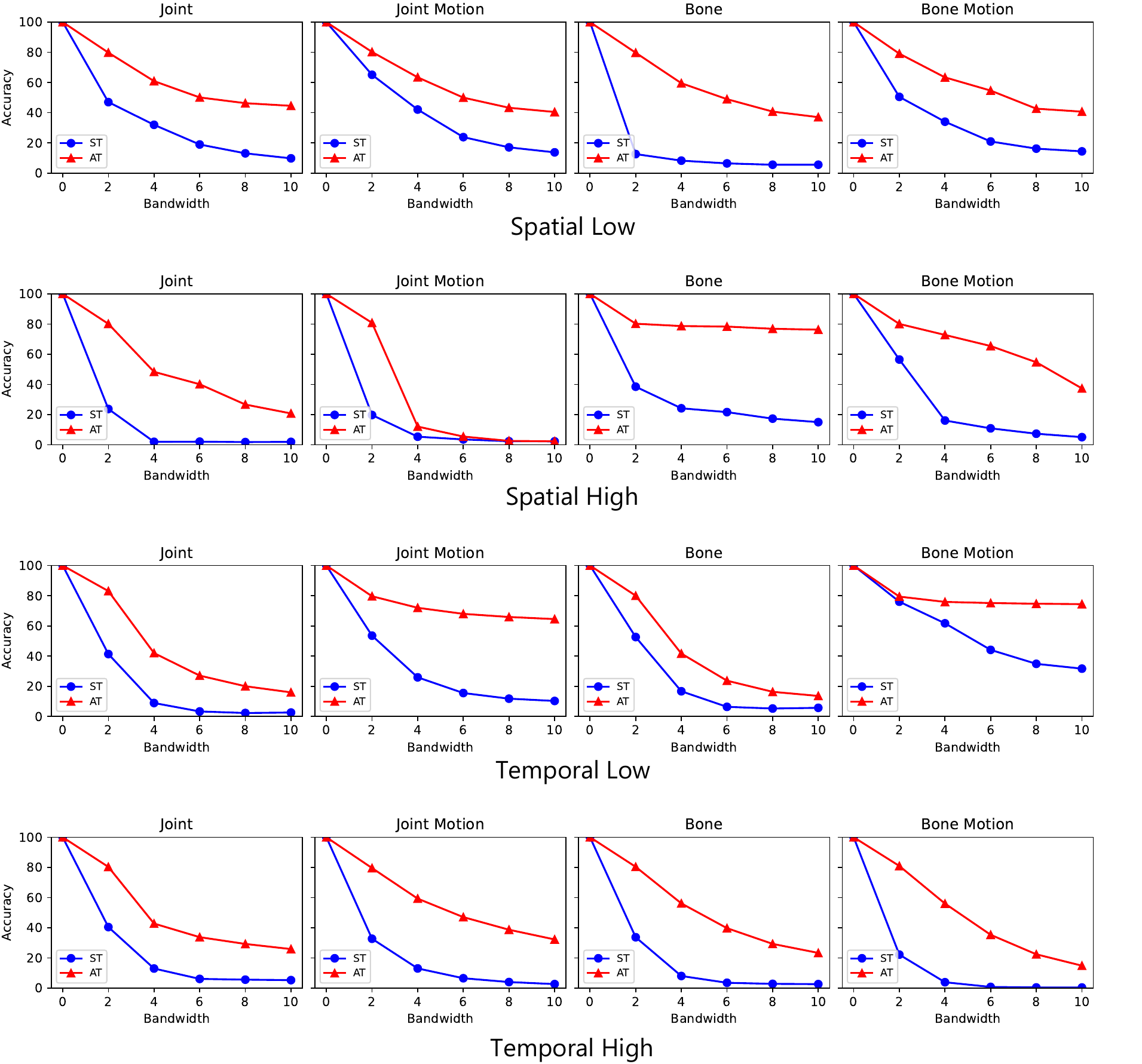}
\caption{
Accuracy comparison of standard-trained (blue) and adversarially-trained (red) GCNs for changes in the
Gaussian noise bandwidth with the scaled norm
in Table.~\ref{tab:scaled_norm1}.
(top) spatial low-pass filter, (second) spatial high-pass
filter, (third) temporal low-pass filter, and (bottom)
temporal high-pass filter.
}
\label{fig:gauss_bandwidth}
\end{figure}

The frequency characteristics of adversarial attacks are examined using the spectral distribution of
successful adversarial attacks.
Fig.~\ref{fig:perturbations} shows
the spectral distributions, which
are obtained by estimating the average amplitude
of successful adversarial examples, i.e.,
$\mathbb{E}\qty[|\mathrm{JFT}(I(\bm{X}+\bm{\delta}_\mathrm{adv})-I(\bm{X}))|]$,
where $\bm{X}$ is clean skeletal data and $\bm{\delta}_\mathrm{adv}$ is an adversarial perturbation that successfully attacks a given GCN.
Here, $I(\cdot)$ is the linear interpolation operation
in Section~\ref{subsec:notations}.

Fig.~\ref{fig:perturbations} shows that
the spectral distributions of the adversarial attacks on the standard-trained models are broadly distributed, whereas those on the adversarially-trained models are concentrated in the lower frequencies.
These frequency characteristics indicate that
the adversarially-trained models capture the features
from the lower frequency signals better than the standard-trained models.
This observation supports the hypothesis
that adversarial training provides robustness against high-frequency perturbations,
as discussed in the previous subsection.
However, the adversarially-trained models with the bone motion feature remain vulnerable in the high-frequency domain (i.e., the rightmost column in Fig.~\ref{fig:perturbations}).
This phenomenon reasonably leads us to conclude that
the bone feature contains high-frequency signals~\cite{abello2021dissecting} as shown in Fig.~\ref{fig:input_spectrum} (i.e., the rightmost column).

\subsubsection{Robustness Trade-off between High-Frequency and Low-Frequency Perturbations}
\label{sec:trade-off}

We examine whether a trade-off exists where adversarially-trained models are robust to high-frequency perturbations
but highly vulnerable to low-frequency perturbations.
This examination is motivated by the existence of the same trade-off as found in CNN-based image classification~\cite{yin2019fourier,chan2022how}.
The trade-off is evaluated by adding low- or high-frequency Gaussian noise, as described in Section~\ref{subsec: DFT_GFT}, to the clean data
that both models correctly classify.
The norm of the Gaussian noise is adjusted such that the accuracy of either the standard-trained model or the adversarially-trained model is approximately 80\%. This adjustment is required for a fair comparison because the appropriate norm differs between the four features. These norms are listed in Table~\ref{tab:scaled_norm1} and denoted as scaled norms. In the following experiments, we use the ST-GCN because there is no significant difference in the frequency characteristics of the two architectures.

\begin{table}[t]
  \caption{List of scaled norms for spatiotemporally filtered Gaussian noises.}
  \centering
  \begin{tabular}{c|c|c|c|c} \hline
    Filter & Joint & Joint Motion & Bone & Bone Motion\\ \hline
    Spatial \& Temporal Low &  78.0 & 5.60 & 22.0 & 1.25\\ \hline
    Spatial \& Temporal High & 355 & 220 & 35.0 & 130\\ \hline
  \end{tabular}
  \label{tab:scaled_norm2}
\end{table}

First, we evaluate the accuracies of the
standard-trained model (blue) and the adversarially-trained model (red), as shown in Fig.~\ref{fig:gauss_norm},
for changes in the Gaussian noise norm while maintaining a
fixed bandwidth of 2.
The norm is changed at 20\%, 40\%, 60\%, 80\%, and
100\% of the scaled norm.
The top two and bottom two rows in Fig.~\ref{fig:gauss_norm}
show the accuracies when perturbed with spatially or temporally filtered Gaussian noises, respectively.
In all cases, the adversarially-trained model (red) is more robust than the standard-trained model (blue).

\begin{figure}[t]
\centering
\includegraphics[width=0.8\columnwidth]{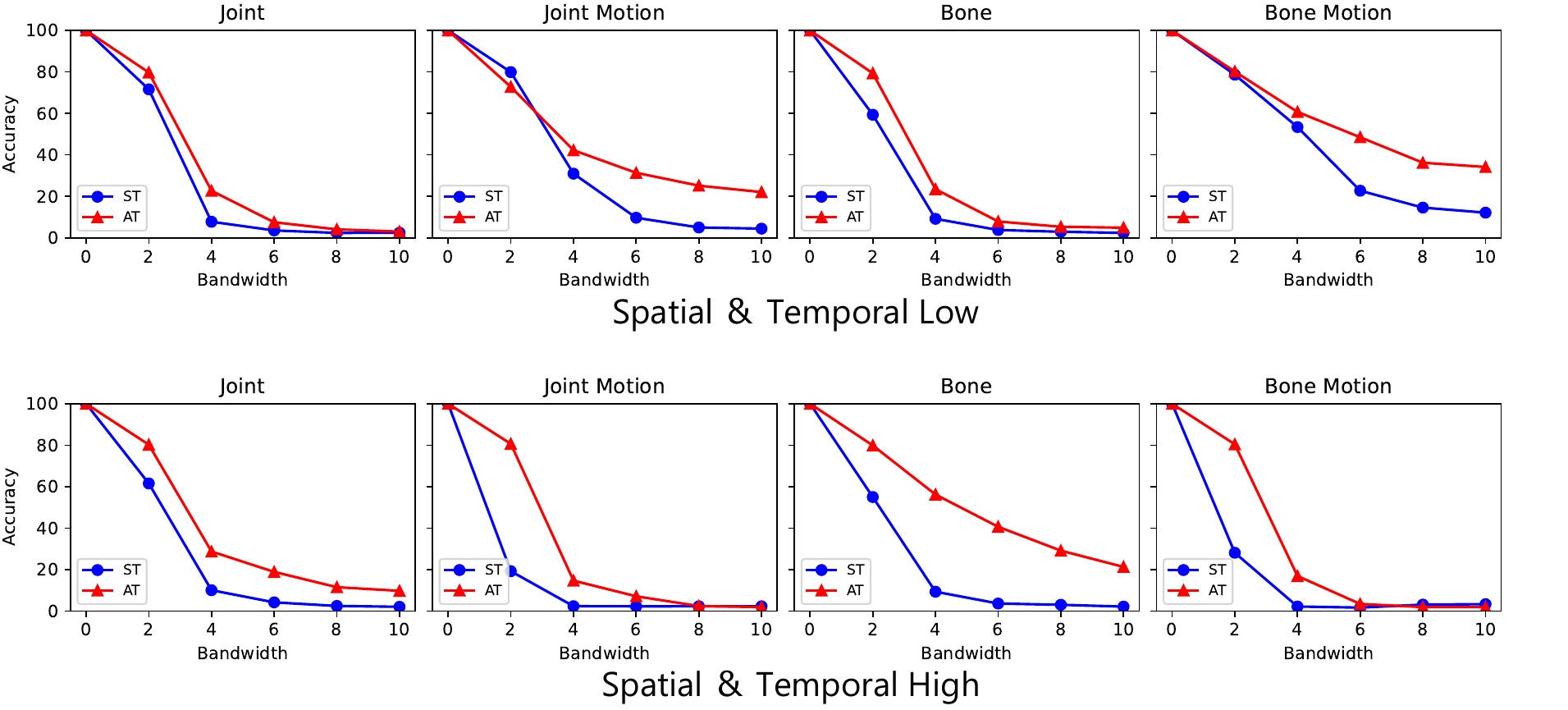}
\caption{Accuracy comparison of standard-trained (blue) and
adversarially-trained (red) GCNs for changes in the
Gaussian noise bandwidth with the scaled norm
in Table~\ref{tab:scaled_norm2}:
(top) spatiotemporal low-pass filter, (bottom)
spatiotemporal high-pass filter.}
\label{fig:gauss_spatial_temporal}
\end{figure}

Next, we evaluate the accuracies of the two models upon changing the Gaussian noise bandwidth
while keeping the norm fixed at the scaled norm.
The results are shown in Fig.~\ref{fig:gauss_bandwidth}
and indicate that the adversarially-trained model (red) is also more robust than the standard model.
Finally,
we evaluate the accuracies
for changes in the spatially and temporally filtered Gaussian noise norm while keeping the norm fixed.
Table~\ref{tab:scaled_norm2} lists the scaled norm
for the experiment, and Fig.~~\ref{fig:gauss_spatial_temporal} also proves
that the adversarially-trained model (red) is again more robust.

In summary, these results indicate that GCN-based skeletal action recognition does not suffer from the same robustness trade-offs as CNNs. 
In some exceptions, the accuracy of the adversarially-trained and standard-trained models are comparable, e.g., the bone motion feature labeled "Temporal Low" in Fig.~\ref{fig:gauss_norm}. 
These results arise because the robustness of the two GCNs does not differ significantly within that frequency band, as evident from the corresponding frequency band of the Fourier heatmap in Fig.~\ref{fig:heatmap}.
Except for such frequency bands,
adversarial training
can improve the GCNs robustness to both low- and high-frequency
perturbations, i.e., without
sacrificing robustness in the low-frequency domain.

\begin{figure}[t]
    \centering
    \includegraphics[width=0.6\linewidth]{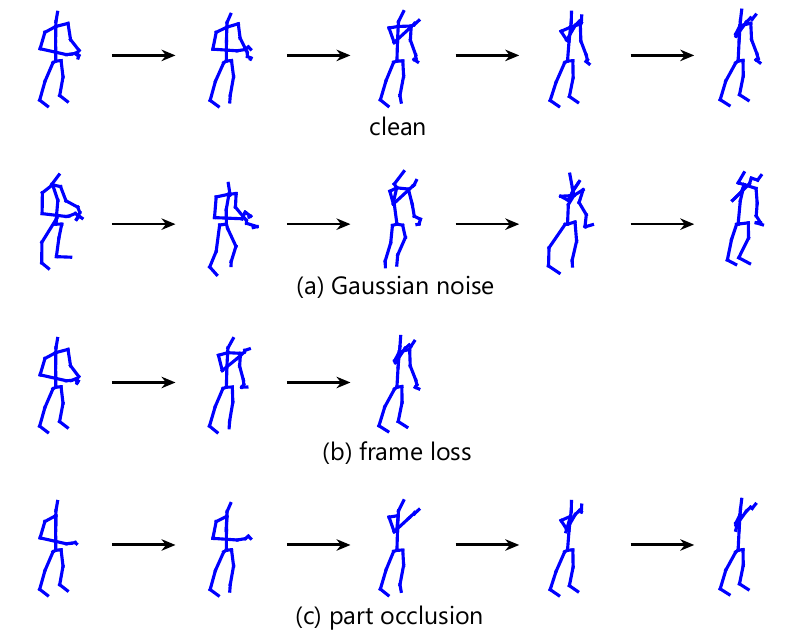}
\caption{Examples of common corruptions in skeleton-based action recognition:
(a) Gaussian noise, (b) frame loss, and (c) part occlusion.}
    \label{fig:corruptions}
\end{figure}

\subsubsection{Robustness to Common Corruptions}
We evaluate the robustness to common corruptions.
In image classification, common corruptions to image data include
high-frequency corruptions, such as Gaussian noise and shot noise,
and low-frequency corruptions, such as motion blur and fog~\cite{hendrycks2018benchmarking}.
The trade-off between accuracy and adversarial robustness makes CNN-based
image classifiers robust to high-frequency corruptions
but vulnerable to low-frequency corruptions.
However, the experiment in the previous section demonstrates that
such a trade-off does not exist for the GCNs.
Therefore, adversarial training is expected to be more robust to common corruptions than standard training, regardless of its frequency spectra. We experimentally examine this conclusion based on the following three common corruptions for skeleton data,
as shown in Fig.~\ref{fig:corruptions}.
\begin{itemize}
\item Gaussian noise: the skeletal data are perturbed
by adding a zero-mean Gaussian noise with standard
deviation $\sigma=0.01, 0.03, 0.05$.
\item Frame loss: Each frame of the skeletal
sequence data is randomly lost.
Loss rate $p$ is a uniform random number in the
$[0, 1]$ interval.
The frame length is adjusted when input to the GCNs
by linearly interpolating the lost frames.
\item Part occlusion~\cite{song2021richly}:
a part of the skeletal data is occluded. The skeletal data are divided into five parts: left arm (part 1), right arm (part 2), both hands (part 3), both legs (part 4), and torso (part 5),
and either part coordinates are set to 0.
\end{itemize}

\begin{figure}[t]
    \centering
    \includegraphics[width=0.8\columnwidth]{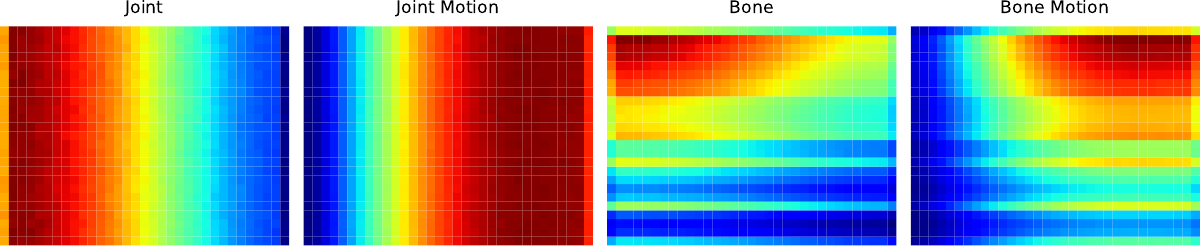}
\caption{Average Fourier spectrum of Gaussian noise corruptions
with standard deviation $\sigma=0.05$ in the spatiotemporal frequency domain.}
    \label{fig:gauss_fourier}
\end{figure}

\begin{figure}[t]
    \centering
    \includegraphics[width=0.8\columnwidth]{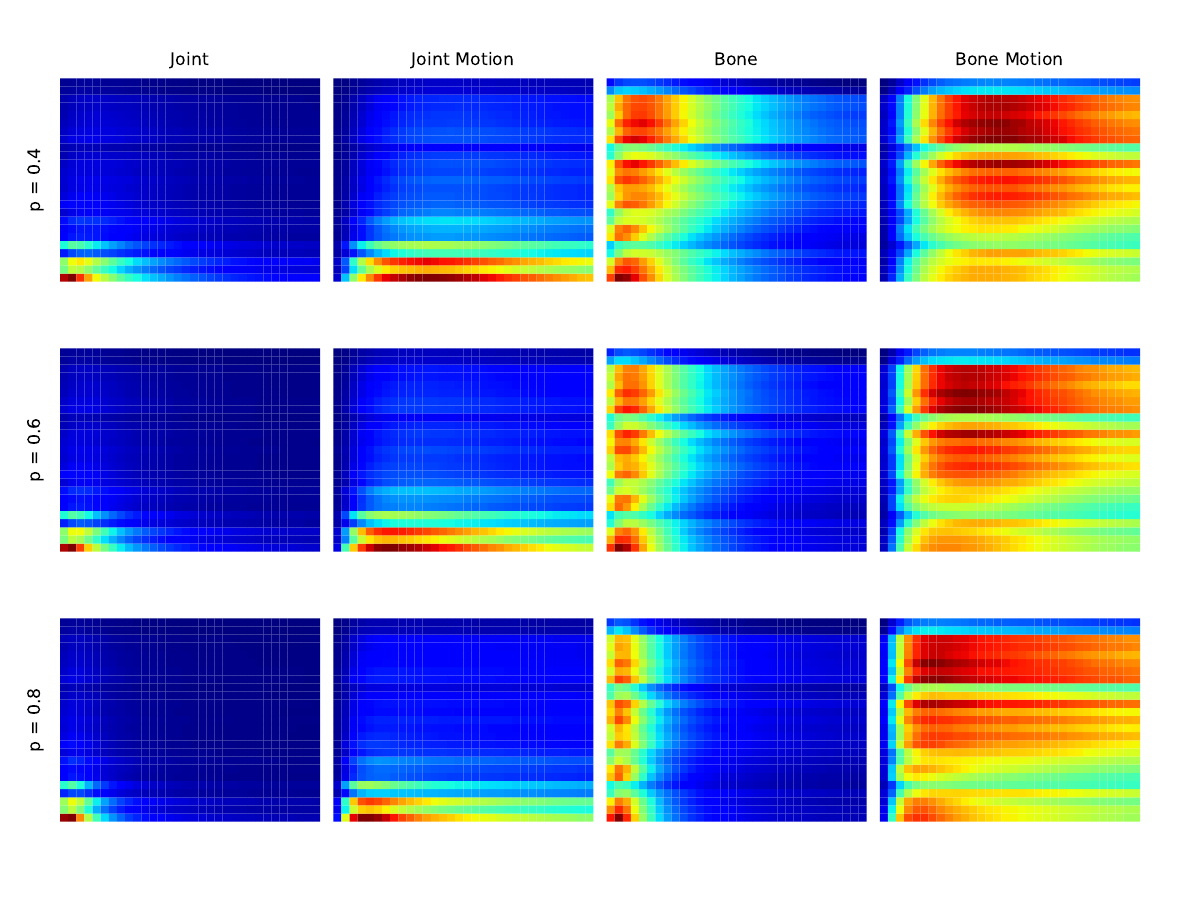}
\caption{
Average Fourier spectrum of frame loss corruptions with
loss rates $p=0.4, 0.6, 0.8$ in the spatiotemporal frequency domain.}
    \label{fig:fl_fourier}
\end{figure}

\begin{figure}[t]
    \centering
    \includegraphics[width=0.8\columnwidth]{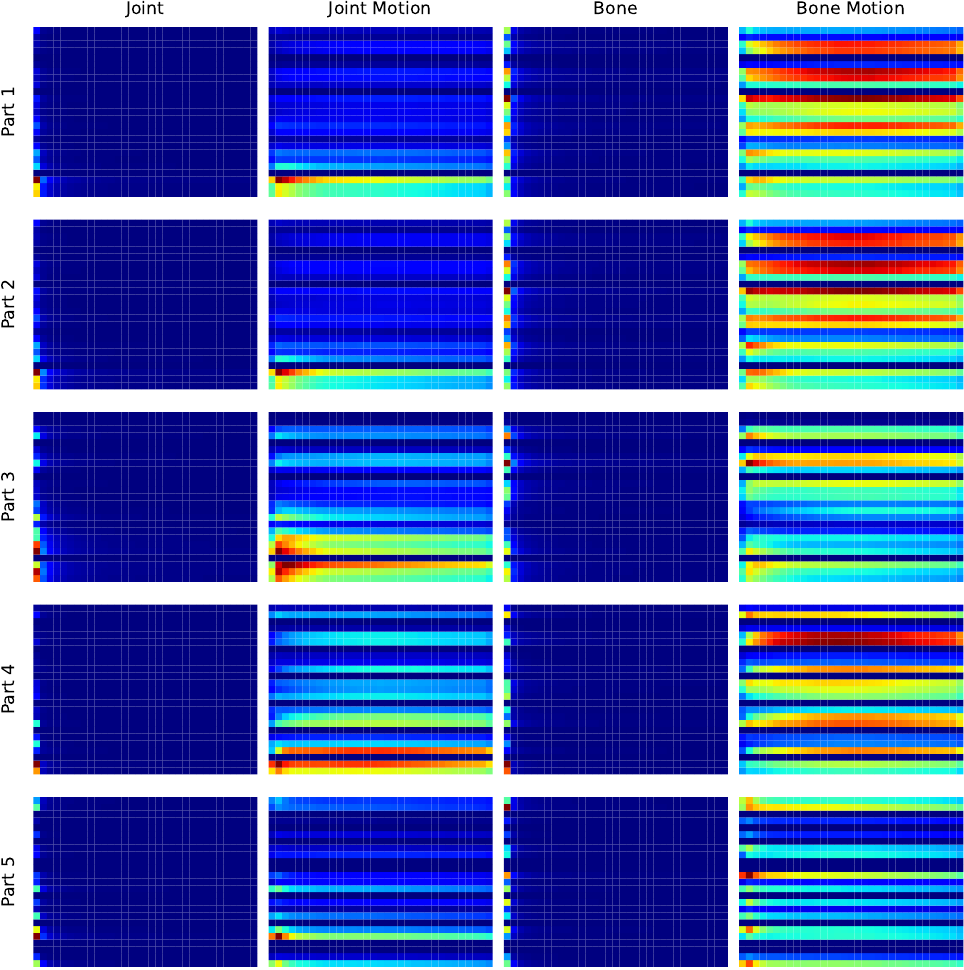}
\caption{Spectral distributions of part occlusion corruptions in the spatiotemporal frequency domain.}
    \label{fig:partocc_fourier}
\end{figure}

To provide a Fourier analysis of the common corruptions, we compute their spectral distributions
as shown in Figs.~\ref{fig:gauss_fourier}--\ref{fig:partocc_fourier}.
These distributions are given by
$\mathbb{E}\qty[|\mathrm{JFT}(I(C(\bm{X}))-I(\bm{X}))|]$,
where $\bm{X}$ is clean skeletal data and $C(\cdot)$ is
one of the three corruptions.

\begin{table}[t]
  \caption{Comparison of accuracy against Gaussian
  noise corruptions with standard deviation $\sigma$ for standard-trained (ST) and adversarially-trained (AT) models.}
  \centering
  \begin{tabular}{c|c|c|c} \hline
    &$\sigma=0.01$ & $\sigma=0.03$ & $\sigma=0.05$  \\ \hline
    ST-GCN ST (Joint) & 94.3\% & 67.4\% & 42.5\% \\ 
    ST-GCN AT (Joint) & \textbf{99.8}\% & \textbf{99.5}\% & \textbf{98.8}\% \\ \hline
    TCA-GCN ST (Joint) & 95.4\% & 82.7\% & 59.8\% \\
    TCA-GCN AT (Joint) & \textbf{99.9}\% & \textbf{99.0}\% & \textbf{96.8}\%  \\ \hline
    ST-GCN ST (Joint Motion) & 74.4\% & 34.1\% & 19.1\% \\
    ST-GCN AT (Joint Motion) & \textbf{98.7}\% & \textbf{87.9}\% & \textbf{64.9}\% \\ \hline
    TCA-GCN ST (Joint Motion) & 74.8\% & 40.2\% & 20.7\%  \\ 
    TCA-GCN AT (Joint Motion) & \textbf{99.2}\% & \textbf{87.6}\% & \textbf{65.7}\% \\ \hline
    ST-GCN ST (Bone) & 87.7\% & 25.6\% & 10.9\%  \\
    ST-GCN AT (Bone) & \textbf{99.8}\% & \textbf{99.0}\% & \textbf{97.1}\%  \\ \hline
    TCA-GCN ST (Bone) & 91.7\% & 50.7\% & 20.1\% \\ 
    TCA-GCN AT (Bone) & \textbf{99.6}\% & \textbf{97.4}\% & \textbf{90.6}\% \\ \hline
    ST-GCN ST (Bone Motion) & 43.5\% & 6.2\% & 1.28\%  \\ 
    ST-GCN AT (Bone Motion) & \textbf{98.1}\% & \textbf{84.0}\% & \textbf{56.5}\% \\ \hline
    TCA-GCN ST (Bone Motion) & 59.6\% & 13.1\% & 2.7\% \\ 
    TCA-GCN AT (Bone Motion) & \textbf{95.8}\% & \textbf{88.7}\% & \textbf{71.9}\% \\ \hline
  \end{tabular}
  \label{tab:gaussian}
\end{table}

\begin{table}[t]
\caption{Comparison of accuracy against frame loss
  corruptions with loss rate $p$ for standard-trained (ST) and adversarially-trained (AT) models.
  }
  \centering
  \begin{tabular}{c|c|c|c} \hline
    &$p = 0.4$ & $p = 0.6$ & $p = 0.8$   \\ \hline
    ST-GCN ST (Joint) & 97.3\% & 94.8\% & 84.4\%  \\
    ST-GCN AT (Joint) & \textbf{97.4}\% & \textbf{95.3}\% & \textbf{88.8}\% \\ \hline
    TCA-GCN ST (Joint) & 96.2\% & 90.5\% & 80.6\%  \\ 
    TCA-GCN AT (Joint) & \textbf{97.7}\% & \textbf{96.2}\% & \textbf{91.1}\% \\ \hline
    ST-GCN ST (Joint Motion) & 92.6\% & 78.8\% & 55.3\% \\ 
    ST-GCN AT (Joint Motion) & \textbf{94.9}\% & \textbf{90.9}\% & \textbf{77.9}\% \\ \hline
    TCA-GCN ST (Joint Motion) & 92.9\% & 81.0\% & 61.3\% \\ 
    TCA-GCN AT (Joint Motion) & \textbf{95.5}\% & \textbf{91.4}\% & \textbf{80.3}\% \\ \hline
    ST-GCN ST (Bone) & \textbf{97.6}\% & \textbf{95.9}\% & 86.7\%\\
    ST-GCN AT (Bone) & 97.5\% & 95.4\% & \textbf{89.2}\% \\ \hline
    TCA-GCN ST (Bone) & 96.1\% & 89.1\% & 78.3\% \\ 
    TCA-GCN AT (Bone) & \textbf{97.7}\% & \textbf{95.6}\% & \textbf{90.7}\% \\ \hline
    ST-GCN ST (Bone Motion) & 93.1\% & 81.3\% & 55.7\%\\
    ST-GCN AT (Bone Motion)  & \textbf{95.9}\% & \textbf{92.3}\% & \textbf{80.1}\% \\ \hline
    TCA-GCN ST (Bone Motion) & \textbf{92.7}\% & \textbf{79.7}\% & 59.6\% \\ 
    TCA-GCN AT (Bone Motion) & 87.9\% & 77.9\% & \textbf{68.8}\%\\ \hline
  \end{tabular}
  \label{tab:frameloss}
\end{table}

\begin{table}[t]
  \caption{Comparison of accuracy against part
  occlusion corruptions for standard-trained (ST) and adversarially-trained (AT) models. }
  \centering
  \begin{tabular}{c|c|c|c|c|c} \hline
    &part 1 & part 2 & part 3 & part 4 & part 5\\ \hline
    ST-GCN ST (Joint) & \textbf{86.7}\% & \textbf{72.8}\% & \textbf{80.5}\% & \textbf{95.5}\% & \textbf{69.3}\% \\
    ST-GCN AT (Joint) & 78.3\% & 61.3\% & 48.5\% & 87.5\% & 63.6\% \\ \hline
    TCA-GCN ST (Joint) & \textbf{84.8}\% &  \textbf{69.6}\% & \textbf{61.8}\% & \textbf{90.4}\% & \textbf{79.4}\% \\ 
    TCA-GCN AT (Joint) & 73.6\% & 58.5\% & 47.7\% & 72.8\% & 74.8\% \\ \hline
    ST-GCN ST (Joint Motion) & \textbf{83.2}\% & \textbf{67.2}\% & \textbf{95.8}\% & \textbf{92.0}\% & 74.1\% \\
    ST-GCN AT (Joint Motion) & 75.0\% & 57.6\% & 62.0\% & 88.2\% & \textbf{89.4}\% \\ \hline
    TCA-GCN ST (Joint Motion) & \textbf{83.5}\% & \textbf{71.2}\% & \textbf{84.1}\% & 85.0\% & 82.7\% \\ 
    TCA-GCN AT (Joint Motion) & 73.5\% & 56.6\% & 68.4\% & \textbf{87.0}\% & \textbf{93.2}\% \\ \hline
    ST-GCN ST (Bone) & \textbf{82.0}\% & \textbf{63.8}\% & 49.4\% & 59.7\% & 46.9\% \\
    ST-GCN AT (Bone) & 65.0\% & 52.0\% & \textbf{62.6}\% & \textbf{75.0}\% & \textbf{54.7}\% \\ \hline
    TCA-GCN ST (Bone) & \textbf{83.8}\% & \textbf{66.0}\% & 33.9\% & \textbf{67.6}\% & 57.6\% \\
    TCA-GCN AT (Bone) & 71.4\% & 52.3\% & \textbf{39.4}\% & 62.9\% & \textbf{60.2}\% \\ \hline
    ST-GCN ST (Bone Motion) & \textbf{84.4}\% & \textbf{67.9}\% & \textbf{85.7}\% & \textbf{88.9}\% & 73.3\% \\
    ST-GCN AT (Bone Motion) & 73.6\% & 56.8\% & 71.1\% & 87.0\% & \textbf{83.8}\% \\ \hline
    TCA-GCN ST (Bone Motion) & \textbf{84.8}\% & \textbf{70.6}\% & \textbf{85.8}\% & 81.4\% & \textbf{83.6}\% \\ 
    TCA-GCN AT (Bone Motion) & 73.2\% & 54.3\% & 58.3\% & \textbf{85.2}\% & 81.7\% \\  \hline
  \end{tabular}
  \label{tab:part_occ}
\end{table}

\paragraph{Gaussian Noise}
Fig.~\ref{fig:gauss_fourier} shows the average Fourier spectrum distributions for the four features, and
Table~\ref{tab:gaussian} lists
the accuracies of the standard-trained and adversarially-trained two GCNs with standard deviation $\sigma=0.01,0.03,0.05$.
As depicted in Fig.~\ref{fig:gauss_fourier}, the frequency characteristics of Gaussian noise corruption differ for the four features.
For example, the joint feature largely includes temporal low-frequency signals (i.e., the left half of the map), whereas the joint motion feature includes temporal high-frequency
signals (i.e., the right half of the map).

For such corruptions, we predict which frequency bands are vulnerable from the Fourier heatmap in Fig.~\ref{fig:heatmap}.
For example, the Fourier heatmap of the joint feature,
as shown in Fig.~\ref{fig:heatmap} (i.e., leftmost column)
indicates that the standard-trained models are vulnerable in the temporal low-frequency domain, but the adversarially-trained
models are more robust in all frequency domains.
Therefore, adversarially-trained models are expected
to provide better accuracies in all cases.
Table~\ref{tab:gaussian} supports this prediction.
The same holds for the other three features.
Hence, adversarial training yields robustness to Gaussian noise corruptions for GCN-based skeletal action recognition.

\paragraph{Frame Loss}
Fig.~\ref{fig:fl_fourier} shows the average Fourier spectra of the frame loss corruptions
for the four features, and
Table~\ref{tab:frameloss} lists
the accuracies of the standard-trained and adversarially-trained two GCNs with frame loss rate $p=0.4,0.6,0.8$.
Table~\ref{tab:frameloss} shows that
the adversarially-trained models achieve higher or comparable
accuracy to the standard-trained models in most cases.
The exception is the case of the TCA-GCN bone motion feature
(bottom row of Table~\ref{tab:frameloss}), where
the standard-trained TCA-GCN provides better performance
in accuracy at loss rates $p=0.4$ and $0.6$. 
This reversal is attributed to the fact that the frame losses at p=0.4 and 0.6 contain more frequencies in the spatial low-frequency and the temporal mid- to high-frequency bands, as shown in 
Fig.~\ref{fig:fl_fourier}. 
In these frequency bands, the Fourier heatmap in Fig.~\ref{fig:heatmap} shows that the standard-trained TCA-GCN is more robust for the bone motion features.
Except for such frequency bands,
the adversarially-trained models
are more robust over a wider frequency range
than the standard-trained models, as with Gaussian corruption.

\paragraph{Part Occlusion}
Finally,
we examine the robustness against part occlusion
corruptions.
Fig.~\ref{fig:partocc_fourier} shows
the average Fourier spectrum distributions,
and Table~\ref{tab:part_occ} lists
the accuracies for each type of model.
In contrast to the other corruptions,
Table~\ref{tab:part_occ} indicates that
the standard-trained models are
more robust in most cases.
In other words, these corruptions cannot be explained using
the Fourier heatmaps.
For example, the spectrum for the
bone motion feature of part 1 (i.e., the top right
in Fig.~\ref{fig:partocc_fourier}) includes
high-frequency spatial signals, and therefore
the adversarially-trained models should be more robust
to the corruption.
However, the results in Table~\ref{tab:part_occ}
demonstrate
that the standard-trained models are more robust to this scenario.
A possible alternative explanation is that the
part occlusion corruptions cause a portion of the
skeletal signals to be missing, resulting in the loss
of the features necessary for action recognition.
The missing signal problem cannot be explained from the Fourier perspective, and other approaches need to be considered.

\section{Conclusions}\label{sec:conclusion}
This study examines the robustness of the GCNs for
skeleton-based action recognition against adversarial
attacks and common corruptions.
Fourier analysis of the 
robustness of GCNs is reported.
We compute the average Fourier spectra of adversarial
perturbations and common corruptions using
the JFT, which is a combination of the GFT and DFT.
Furthermore, the frequency effects of both standard and adversarial training
are explored using the Fourier heatmap.

Our experiments reveal that the standard-trained models are sensitive to high-frequency perturbations,
and adversarial training suppresses this sensitivity and enhances robustness to high-frequency perturbations.
This robustness against high frequencies is also known in CNN-based image classification.
Our interesting finding is that the GCNs for skeleton-based
action recognition do not suffer from a robustness trade-off between 
adversarial robustness and low-frequency perturbations (as do CNNs). 
While the absence of the trade-off remains an open problem, our experiments lead to the following observations.
As depicted in Fig.~\ref{fig:input_spectrum}, all four skeletal features display distinct frequency characteristics, yet no trade-off is observed. This is true even for the joint feature, which exhibits a low-frequency concentration similar to natural images. Thus, our experimental results suggest that the
discrepancy in robustness primarily stems from the inherent differences between CNNs and GCNs, rather than the frequency characteristics of the skeletal features.
It will be necessary to explore this difference not only through Fourier analysis but also through other analysis methods and perspectives.
Furthermore, given the transition from CNNs and GCNs to Transformer architectures in computer vision, frequency analysis of Transformers, as in \cite{tancik2020fourier}, will be important research for robust computer vision.

\bibliographystyle{unsrt}  
\bibliography{bib}

\end{document}